%% file: ICC2013.tex
\documentclass[conference]{IEEEtran}
\usepackage{cite}

\ifCLASSINFOpdf
   \usepackage[pdftex]{graphicx}
  \DeclareGraphicsExtensions{.pdf,.jpeg,.png}
\else
  \usepackage[dvips]{graphicx}
  \DeclareGraphicsExtensions{.eps}
\fi
\usepackage[cmex10]{amsmath}
\usepackage{amssymb}
\usepackage{dsfont}

\usepackage{mdwmath}
\usepackage{mdwtab}

\usepackage{pstricks,pst-plot,pst-node,setspace}
\usepackage{t1enc}
\usepackage{psfrag}

\usepackage[tight,footnotesize]{subfigure}
\usepackage{url}

\usepackage{comment}

\newcommand{\vect}[1]{ \boldsymbol{#1} }
\newcommand{\matr}[1]{ \boldsymbol{#1} }

\newcommand{\hvect}[1]{ \widehat{\boldsymbol{#1}} }

\newcommand{\Var}{\mathbb{V}\text{ar}}
\newcommand{\CGaussPDF}{\mathrm{CN}}

\newcommand{\GammaPDF}{\mathrm{Ga}}
\newcommand{\ipilot}{\mathcal{P}}
\newcommand{\compnum}{\mathbb{C}}

\DeclareMathOperator*{\diag}{diag}

\newcommand{\trans}{\mathrm{T}}
\newcommand{\hermit}{\mathrm{H}}

\IEEEoverridecommandlockouts
\begin{document}
%
\title{A Fast Iterative Bayesian Inference Algorithm\\for Sparse Channel Estimation}

\author{\IEEEauthorblockN{Niels Lovmand Pedersen, 
Carles Navarro Manch\'{o}n 
and Bernard Henri Fleury}
\IEEEauthorblockA{Department of Electronic Systems, Aalborg University\\
Niels Jernes Vej 12, DK-9220 Aalborg, Denmark, Email: \{nlp,cnm,bfl\}@es.aau.dk} 
\thanks{(c) 2013 IEEE. Personal use of this material is permitted. Permission from IEEE must be obtained for all other users, including reprinting/republishing this material for advertising or promotional purposes, creating new collective works for resale or redistribution to servers or lists, or reuse of any copyrighted components of this work in other works.}
}


\maketitle

\begin{abstract}
\input{abstract.tex}
\end{abstract}


%
\IEEEpeerreviewmaketitle

\section{Introduction}
\input{introduction.tex}

\section{System Description \label{sec:system}}
\input{signalmodel.tex}

\section{Bayesian Inference Learning \label{sec:bil}}
\input{bil.tex}

\section{Numerical Results \label{sec:simulations}}
\input{simulations.tex}

\section{Conclusion}
\input{conclusion.tex}

\section*{Acknowledgment}

This work was supported in part by the 4GMCT cooperative research project, funded by Intel Mobile Communications, Agilent Technologies, Aalborg University and the Danish National Advanced Technology Foundation, and by the project ICT-248894 Wireless Hybrid Enhanced Mobile Radio Estimators (WHERE2). 

\IEEEtriggeratref{14}


%

\bibliographystyle{IEEEtran}
\bibliography{Hcomplex}

\end{document}

%% file: abstract.tex
In this paper, we present a Bayesian channel estimation algorithm for multicarrier receivers based on pilot symbol observations. The inherent sparse nature of wireless multipath channels is exploited by modeling the prior distribution of multipath components' gains with a hierarchical representation of the Bessel K probability density function; a highly efficient, fast iterative Bayesian inference method is then applied to the proposed model. The resulting estimator outperforms other state-of-the-art Bayesian and non-Bayesian estimators, either by yielding lower mean squared estimation error or by attaining the same accuracy with improved convergence rate, as shown in our numerical evaluation.   

%% file: introduction.tex
The accuracy of channel estimation is a crucial factor determining the overall performance in wireless communication systems and networks, in terms of bit-error-rate (BER) and throughput but also of location accuracy when these systems are equipped with positioning capabilities. 
When the underlying structure of the channel responses to be estimated is sparse, compressive sensing and sparse signal representation can be very powerful tools for the design of channel estimators. 

Compressive sensing techniques have attracted considerable attention in recent years due to their ability to be incorporated in a wide range of applications. Typically, the signal model considered reads
\begin{align}
\vect{y} = \matr{\Phi}\vect{\alpha} + \vect{w}
\label{eq:model}
\end{align}
where $\vect{y}\in\compnum^{M\times1}$ is the measurement vector and $\matr{\Phi} = [\vect{\phi}_1,\ldots,\vect{\phi}_L] \in\compnum^{M\times L}$ is the known dictionary matrix with $L>M$ column vectors $\vect{\phi}_l$, $l=1,\ldots,L$. The vector $\vect{w} \in \compnum^{M\times1}$ represents the samples of additive white Gaussian noise with covariance matrix $\lambda^{-1}\matr{I}$ and precision parameter $\lambda > 0 $. Finally, $\vect{\alpha} = [\alpha_1,\dots,\alpha_L]^\trans\in\compnum^{L\times1}$ is the vector of weights whose entries are mostly zero. By obtaining a sparse estimate of $\vect{\alpha}$ we can accurately represent $\matr{\Phi}\vect{\alpha}$ with a minimal number of column vectors in $\matr{\Phi}$. 

In the literature many Bayesian and non-Bayesian methods have been proposed for sparse signal representation. The latter methods include the very popular convex optimization based methods for LASSO regression \cite{Tibshirani1994,Chen1998} and greedy constructive algorithms such as orthogonal matching pursuit (OMP) \cite{Tropp2004} and compressive sampling MP (CoSaMP) \cite{Needell2009}. In sparse Bayesian learning (SBL) \cite{Tipping2001,WipfRao04}, a prior probability density function (pdf) $p(\vect{\alpha})$ is specified so that a sparse estimate $\hvect{\alpha}$ is obtained. A widely applied SBL algorithm is the relevance vector machine (RVM) \cite{Tipping2001}, where a hierarchical representation\footnote{The hierarchical representation involves specifying a conditional prior pdf $p(\vect{\alpha}|\vect{\gamma})$ and a hyperprior pdf $p(\vect{\gamma})$.} of the student-t pdf is used for the prior pdf $p(\vect{\alpha})$. An EM algorithm is then derived based on this prior model for the estimation of the weights. Similarly, \cite{Figueiredo} uses the EM algorithm based on a hierarchical representation of the Laplace pdf.\footnote{Note that the hierarchical representation of the Laplace pdf used in \cite{Figueiredo} and \cite{Babacan2010} is only valid for real-valued variables. In \cite{Pedersen2012a}, we extend this representation to cover complex-valued variables as well.} This algorithm can be seen as the Bayesian version of the LASSO estimator. 
Though the sparse Bayesian inference algorithms proposed in \cite{Tipping2001} and \cite{Figueiredo} are guaranteed to converge, they are also known to suffer from high computational complexity and low convergence rate - many iterations are needed before they terminate. To circumvent this, a fast Bayesian inference algorithm, known as Fast-RVM, is proposed in \cite{Tipping2003}. Following this approach, the Fast-Laplace algorithm is formulated in \cite{Babacan2010}. However, even though the algorithms in \cite{Tipping2003} and \cite{Babacan2010} do lead to faster convergence than their EM counterparts in \cite{Tipping2001} and \cite{Figueiredo}, they still suffer from slow convergence especially in low and moderate signal-to-noise ratio (SNR) regimes as we show in this paper.

The estimation of the wireless channel is a practical example where compressive sensing techniques are utilized. The reason is that the response of the wireless channel typically holds a few dominant multipath components and therefore has the characteristic of being sparse \cite{Bajwa2010}. When sparse channel models are assumed it seems natural to use tools available from compressive sensing and sparse signal representation to estimate the parameters of said channel models. LASSO regression, OMP, and CoSaMP have been widely applied to the problem of pilot-assisted channel estimation in orthogonal frequency-division multiplexing (OFDM), cf., \cite{Berger2010,Huang2010,Taubock2010}. Bayesian methods have also been previously proposed for wireless communication systems. Examples include the estimation of the dominant multipath components in the response of wireless channels \cite{ShutinFleuryVBSAGEChannel} and joint channel estimation and decoding for clustered sparse channels \cite{Schniter2011}. In \cite{Pedersen2012}, we have proposed a variational Bayesian inference algorithm for the estimation of the wireless channel in OFDM. The resulting estimator, however, suffers from the same complexity and convergence rate issues as those in \cite{Tipping2001} and \cite{Figueiredo}.

In this paper, we present a fast iterative sparse Bayesian estimation algorithm for pilot-assisted channel estimation in OFDM wireless receivers. 
We follow the fast inference framework outlined in \cite{Tipping2003} based on the hierarchical prior model of the Bessel K pdf for sparse estimation that we propose in \cite{Pedersen2012a,Pedersen2012}. Our estimator drastically increases the convergence speed compared to similar algorithms such as Fast-RVM and Fast-Laplace with no penalization in performance and achieves favorable BER and mean-squared error (MSE) performance as compared to both Bayesian and non-Bayesian state-of-the-art methods.

%% file: signalmodel.tex
\subsection{OFDM Signal Model}
We consider a single-input single-output OFDM system with $N$ subcarriers. A cyclic prefix (CP) is added to eliminate inter-symbol interference between consecutive OFDM blocks  and the channel response is assumed static during the transmission of each OFDM block. The received baseband signal $\vect{r} \in \compnum^N$ for a given OFDM block reads
\begin{align}
	\vect{r} = \matr{X}\vect{h}+\matr{n}.
	\label{eq:ofdm}
\end{align}
The diagonal matrix $\matr{X} = \diag ( x_1, x_2, \ldots , x_N )$ contains the complex-modulated symbols. The entries in $\vect{h} \in \compnum^{N}$ are the samples of the channel frequency response at all $N$ subcarriers. Finally, $\vect{n} \in \compnum^N $ is a zero-mean complex symmetric Gaussian random vector whose entries are independent with variance $\lambda^{-1}$.

Let the pilot pattern be characterized by the set $\ipilot \subseteq \{1,\ldots,N\}$ containing the indices of subcarriers reserved for pilot transmission. The received signals observed at the pilot positions $\vect{r}_\ipilot = [r_n : n \in \ipilot]^\trans$ are then divided each by their corresponding pilot symbol in $\matr{X}_\ipilot = \diag ( x_n : n \in \ipilot )$ to produce the vector of observations 
\begin{align}
	\vect{y} \triangleq (\matr{X}_\ipilot)^{-1}\vect{r}_{\ipilot} = \vect{h}_{\ipilot} + (\matr{X}_\ipilot)^{-1}\vect{n}_{\ipilot}
	\label{eq:pilotobs}
\end{align} 
where $\vect{h}_{\ipilot}$ and $\vect{n}_{\ipilot}$ are defined analogously to $\vect{r}_\ipilot$. We assume that all $M \triangleq |\ipilot| <N$ pilot symbols hold unit power so that the statistics of the noise term $(\matr{X}_\ipilot)^{-1}\vect{n}_\ipilot$ remain unchanged. 


We consider a frequency-selective, block-fading wireless channel
with impulse response modeled as a sum of multipath components:
\begin{align}
	g(\tau) = \sum_{k=1}^{K} \beta_k\delta\left(\tau - \tau_k\right).
	\label{eq:channel}
\end{align}
In this expression, $\beta_k$ and $\tau_k$ are respectively the complex weight and the (continuous) delay of the $k$th multipath component, $K$ is the total number of multipath components, and $\delta(\cdot)$ is the Dirac delta function. The channel parameters $\beta_k$, $\tau_k$, and $K$ are all random variables and may vary from the transmission of one OFDM block to the next. Additional details regarding the assumptions on the channel model are provided in Section~\ref{sec:simulations}. 


By using the parametric model \eqref{eq:channel} of the channel, we can rewrite \eqref{eq:pilotobs} as 
\begin{align}
	\vect{y} = \matr{T}(\vect{\tau})\vect{\beta} + \vect{w}
	\label{eq:yobs}
\end{align}
with $\vect{h}_\ipilot = \matr{T}(\vect{\tau})\vect{\beta}$, $\vect{w} = (\matr{X}_\ipilot)^{-1}\vect{n}_\ipilot$, $\vect{\beta} = \left[\beta_1,\ldots,\beta_K\right]^\trans$, $\vect{\tau} = \left[\tau_1,\ldots,\tau_K\right]^\trans$, and $\matr{T}(\vect{\tau}) \in \compnum^{M\times K}$ with entries 
\begin{align}
	\matr{T}(\vect{\tau})_{m,k} \triangleq \exp\left(-j2\pi f_m \tau_k\right), \begin{array}{l}  m = 1,2,\ldots,M \\ \;k = 1,2,\ldots,K \end{array}
	\label{eq:Measindices}
\end{align}
where $f_m$ denotes the frequency of the $m$th pilot subcarrier.

\subsection{Compressive Sensing Signal Model}

In order to apply sparse representation methods for the estimation of $\vect{h}$ in \eqref{eq:ofdm}, we must first recast the signal model in \eqref{eq:yobs} into the form of \eqref{eq:model}. The main limitation to do so is that the delay entries in $\vect{\tau}$ are, a priori, unknown at the receiver. To circumvent this, we consider a grid of uniformly-spaced delay samples in the interval $[0,\tau_{\mathrm{max}}]$:
\begin{align}
	\vect{\tau}_d = \Big[0,\frac{T_\mathrm{s} }{\zeta},\frac{2T_\mathrm{s} }{\zeta}, \ldots, \tau_{\mathrm{max}} \Big]^\trans
\label{eq:delay_sampled}
\end{align}
with $\zeta>0$ such that $\zeta \tau_{\mathrm{max}} / T_\mathrm{s}$ is an integer. The symbols $\tau_{\mathrm{max}}$ and $T_\mathrm{s}$ denote respectively the maximum excess delay of the channel and the sampling time. The dictionary matrix $\matr{\Phi} \in \compnum^{M\times L}$ is now defined as $\matr{\Phi} = \matr{T}(\vect{\tau}_d)$. Thus, the entries of $\matr{\Phi}$ are of the form \eqref{eq:Measindices} with argument $\vect{\tau}_d$. The number of columns $L = \zeta \tau_{\mathrm{max}} / T_\mathrm{s} + 1$ in $\matr{\Phi}$ is thereby inversely proportional to the selected delay resolution $T_\mathrm{s} /\zeta$. The selection of $\vect{\tau}_d$ impacts the dimension of $\vect{\alpha}$. By assuming a vector $\vect{\alpha}$ with many more entries than the number of multipath components, we expect most of the entries in $\vect{\alpha}$ to be zero. Therefore, we use compressive sensing techniques to obtain sparse estimates of $\vect{\alpha}$.     

Notice that the signal model \eqref{eq:model} with $\matr{\Phi} = \matr{T}(\vect{\tau}_d)$ is an approximation of the true signal model \eqref{eq:yobs}. The estimate of the channel vector at the pilot subcarriers is then $\hvect{h}_\ipilot = \matr{\Phi}\hvect{\alpha}$. In order to estimate the full channel $\vect{h}$ in \eqref{eq:ofdm} the dictionary $\vect{\Phi}$ is appropriately expanded to include a row corresponding to each of the $N$ subcarrier frequencies. Thus, $\hvect{h} = \matr{\Phi}^{\mathrm{full}}\hvect{\alpha}$ with  
\begin{align}
	\matr{\Phi}^{\mathrm{full}}_{n,l} \triangleq \exp\left(-j2\pi f_n \tau_{d_l}\right), \begin{array}{l}  n = 1,2,\ldots,N \\ \;l = 1,2,\ldots,L \end{array}
	\label{eq:Dictfull}
\end{align}
where $f_n$ denotes the frequency of the $n$th subcarrier.

%% file: bil.tex
We now present the iterative sparse Bayesian inference algorithm for channel estimation proposed in this paper. First, we detail the hierarchical prior model leading to the Bessel K pdf for each entry of $\vect{\alpha}$. Based on this model, we apply a fast Bayesian algorithm to estimate the unknown model parameters. 
Finally, we briefly comment on the relationship between our algorithm and other similar state-of-the-art approaches.

\subsection{The Probabilistic Model}

Instead of working directly with the prior pdf $p(\vect{\alpha})$, in the SBL framework, $p(\vect{\alpha})$ is usually modeled using a two-layer hierarchical prior model involving a conditional prior pdf $p(\vect{\alpha}|\vect{\gamma})$ and a hyperprior pdf $p(\vect{\gamma})$. With this design, the resulting probabilistic model for signal model \eqref{eq:model} is given by
\begin{align}
p(\vect{y},\vect{\alpha},\vect{\gamma},\lambda) &= p(\vect{y}|\vect{\alpha},\lambda)p(\lambda)p(\vect{\alpha}|\vect{\gamma})p(\vect{\gamma}) \notag \\
&=p(\vect{y}|\vect{\alpha},\lambda)p(\lambda)\prod_{l=1}^L p(\alpha_l|\gamma_l)p(\gamma_l).
	\label{eq:jointpdf2layer}
\end{align}
Due to \eqref{eq:model}, $p(\vect{y}|\vect{\alpha},\lambda)$ is multivariate Gaussian: $p(\vect{y}|\vect{\alpha},\lambda)=\CGaussPDF(\vect{y}|\matr{\Phi}\vect{\alpha},\lambda^{-1}\matr{I})$.\footnote{Here, $\CGaussPDF(\cdot|\vect{a},\matr{B})$ denotes a complex Gaussian pdf with mean vector $\vect{a}$ and covariance matrix $\matr{B}$. We shall also make use of $\GammaPDF(\cdot|a,b)= \frac{b^a}{\Gamma(a)}x^{a-1}\exp(-bx)$, which denotes a gamma pdf with shape parameter $a$ and rate parameter $b$.} For the noise precision $\lambda$, we select a constant prior, i.e., $p(\lambda) \propto 1$. 

The design of the factors $p(\alpha_l|\gamma_l)$ and $p(\gamma_l)$ for each weight $\alpha_l$ heavily influences the sparsity-inducing property of the prior model. We adopt the hierarchical structure of the Bessel K pdf, where the first layer is defined as $p(\alpha_l|\gamma_l) = \CGaussPDF(\alpha_l|0,\gamma_l)$ and the second layer is selected to be  $p(\gamma_l) = \GammaPDF(\gamma_l|\epsilon,\eta)$. With these choices, we compute the marginal pdf
\begin{align}
p(\alpha_l;\epsilon,\eta)= \frac{2\eta^{\frac{\epsilon+1}{2}}}{\pi \Gamma(\epsilon)}|\alpha_l|^{\epsilon -1}K_{\epsilon-1}(2\sqrt{\eta}|\alpha_l|).
\label{eq:palpha}
\end{align}
In this expression, $K_\nu(\cdot)$ is the modified Bessel function of the second kind and order $\nu \in \mathds{R}$. The parameter $\epsilon$ determines the sparsity-inducing property of the Bessel K pdf \cite{Pedersen2012a}. The selection $\epsilon=0$ greatly enforces sparseness on the estimate as more probability mass concentrates around the origin. As a consequence, the mode of the resulting posterior pdf $p(\vect{\alpha}|\vect{y},\epsilon,\eta)$ is more likely to be found close to the axes. However, selecting a too high $\epsilon$ ($\epsilon\geq1$) may lead to over-fitting and thereby non-sparse results. Thus, this parameter has a similar functionality as the parameter $p$ in the FOCUSS algorithm \cite{Gorodnitsky1997}.     

\subsection{Fast Iterative Bayesian Inference \label{sec:fast_besselk}}

Given fixed estimates $\hat{\vect{\gamma}}$ and $\hat{\lambda}$, the posterior pdf $p(\vect{\alpha}|\vect{y},\hat{\vect{\gamma}},\hat{\lambda})$ is a multivariate Gaussian, i.e., $p(\vect{\alpha}|\vect{y},\hat{\vect{\gamma}},\hat{\lambda})=\CGaussPDF\left( \vect{\alpha}|\hat{\vect{\mu}}, \widehat{\matr{\Sigma}} \right)$ with
\begin{align}
	\widehat{\matr{\Sigma}} &= \left(\hat{\lambda} \matr{\Phi}^\hermit\matr{\Phi}+ \widehat{\matr{\Gamma}}^{-1}\right)^{-1}, \label{eq:alpha_cov}\\
	\hat{\vect{\mu}} &= \hat{\lambda} \widehat{\matr{\Sigma}}\matr{\Phi}^\hermit\vect{y}  \label{eq:alpha_mu}
\end{align}
where $\widehat{\matr{\Gamma}}=\diag (\hat{\gamma}_1,\ldots,\hat{\gamma}_L )$. The hyperparameters $\vect{\gamma}$ and $\lambda$ are estimated by maximizing \cite{Tipping2001,WipfRao04}
\begin{align}
	\mathcal{L}(\vect{\gamma},\lambda)
	= \log ( p(\vect{y}|\vect{\gamma},\lambda)p(\vect{\gamma})p(\lambda) ).
	\label{eq:sbl_cost}
\end{align} 
The cost function \eqref{eq:sbl_cost} can be iteratively maximized using the EM algorithm by noting that $\vect{\alpha}$ and $\vect{y}$ are complete data for $\vect{\gamma}$ and $\lambda$. Following the classical EM formulation, the E-step equivalently computes  \eqref{eq:alpha_cov}-\eqref{eq:alpha_mu} and the M-step computes
\begin{align}
	\hat{\gamma}_l &= \frac{(\epsilon-2) + \sqrt{(\epsilon-2)^2+4\eta\langle|\alpha_l|^2\rangle}}{2\eta},\;\; l = 1,\ldots,L, \label{eq:gamma_mode} \\
		\hat{\lambda} &= \frac{M}{\langle \|\vect{y}-\matr{\Phi}\vect{\alpha}\|_2^2 \rangle} \label{eq:lambda_mode}. 
\end{align}
The expectation $\langle\cdot\rangle$ in the above expressions are evaluated with respect to the posterior pdf $p(\vect{\alpha}|\vect{y},\hat{\vect{\gamma}},\hat{\lambda})$, where $\hat{\vect{\gamma}}$ and $\hat{\lambda}$ are the estimates computed in the previous iteration. After an initialization procedure, the individual quantities in \eqref{eq:alpha_cov}--\eqref{eq:alpha_mu} and \eqref{eq:gamma_mode}--\eqref{eq:lambda_mode} are iteratively updated until convergence.

The above EM algorithm suffers from two main disadvantages: high computational complexity of the update \eqref{eq:alpha_cov} and low rate of convergence. 
In order to overcome the first drawback a greedy procedure as in \cite{Tipping2003} can be adopted: 
as most of the entries in $\vect{\vect{\alpha}}$ are mostly zero, one may start out with an ``empty'' dictionary matrix and incrementally fill the dictionary by adding column vectors. 
To circumvent the drawback of low convergence rate, we compute the stationary points of the EM update $\hat{\gamma}_l$ in \eqref{eq:gamma_mode}. For this, we fix $\hat{\gamma}_k$, $k\neq l$ at their current estimates, while computing a sequence of estimates $\{\hat{\gamma}^{[t]}_l\}_{t=1}^T$ according to \eqref{eq:gamma_mode} for $T\rightarrow\infty$.\footnote{Notice that $\langle|\alpha_l|^2\rangle$ in \eqref{eq:gamma_mode} is a function of $\hat{\gamma}_l$ as seen from \eqref{eq:alpha_cov} and \eqref{eq:alpha_mu}.} In this way, we update the estimates of the components in $\{\hat{\gamma}_1,\ldots,\hat{\gamma}_N\}$ sequentially, instead of jointly. The generalized EM framework justifies this modification. As shown in \cite{Pedersen2012a}, $\hat{\gamma}^{[\infty]}_l$ corresponds in fact to the (local) extrema of 
\begin{align}
\ell(\gamma_l) &= \mathcal{L}(\gamma_l,\hat{\vect{\gamma}}_{-l},\hat{\lambda}) = -\log|1+\gamma_ls_l|\notag\\
&\;\; + \frac{|q_l|^2}{\gamma_l^{-1}+s_l} + (\epsilon-1) \log \gamma_l - \eta\gamma_l+c
\end{align}
with $c$ being a constant encompassing the terms independent of $\gamma_l$ and the definitions $s_l \triangleq \vect{\phi}_l^\hermit\matr{C}^{-1}_{-l}\vect{\phi}_l$, $q_l \triangleq \vect{y}^\hermit\matr{C}^{-1}_{-l}\vect{\phi}_l$, and $\mathbf{C}=\hat{\lambda}^{-1}\matr{I}+\sum_{k\neq l}\hat{\gamma}_k\vect{\phi}_k\vect{\phi}_k^\hermit + \gamma_l\vect{\phi}_l\vect{\phi}_l^\hermit = \matr{C}_{-l} + \gamma_l\vect{\phi}_l\vect{\phi}_l^\hermit $.\footnote{For the derivation of $\ell(\gamma_l)$, we exploit that $p(\vect{y}|\vect{\gamma},\hat{\lambda})$ is Gaussian with mean zero and covariance matrix $\mathbf{C} = \hat{\lambda}^{-1}\matr{I}+\matr{\Phi}\matr{\Gamma}\matr{\Phi}^\hermit$.} Note that the definition domain of $\ell(\gamma_l)$ is $\mathbb{R}^+$. Now, taking the derivative of $\ell(\gamma_l)$ with respect to $\gamma_l$ and equating the result to zero yields the cubic equation
\begin{align}
	0 
	&=\eta s_l^2\gamma_l^3 + \gamma_l^2[ 2\eta s_l - (\epsilon-2)s_l^2 ] \notag \\
	&\;\;+ \gamma_l [ \eta + (3-2\epsilon)s_l - |q_l|^2 ] - (\epsilon-1).
	\label{eq:cubic}
\end{align}
In general \eqref{eq:cubic} has three solutions when $\gamma_l$ ranges through $\mathbb{R}$. These can be determined analytically with a feasible solution for $\gamma_l$ constrained to be positive. The analysis of the sparsity-inducing property of the Bessel K pdf in \cite{Pedersen2012a} shows that we should select $\epsilon$ small. 
When $\epsilon<1$, \eqref{eq:cubic} has at least one negative solution as $-(\epsilon-1) > 0$. Therefore, \eqref{eq:cubic} has either no real positive solution or two real positive solutions $\hat{\gamma}_l^{(i)}$ and $\hat{\gamma}_l^{(ii)}$. In the former case, no feasible solution to $\ell(\gamma_l)$ exists and the corresponding column vector $\vect{\phi}_l$ is not added to the dictionary. In the latter case, we simply select $\hat{\gamma}_l^{(i)}$ if $\ell(\hat{\gamma}_l^{(i)}) > \ell(\hat{\gamma}_l^{(ii)})$ and $\hat{\gamma}_l^{(ii)}$ otherwise.     

\begin{figure*}[!t]
\centering
\centerline{
\subfigure[\label{fig:BER_EbN0}]{\includegraphics[width=0.3\linewidth]{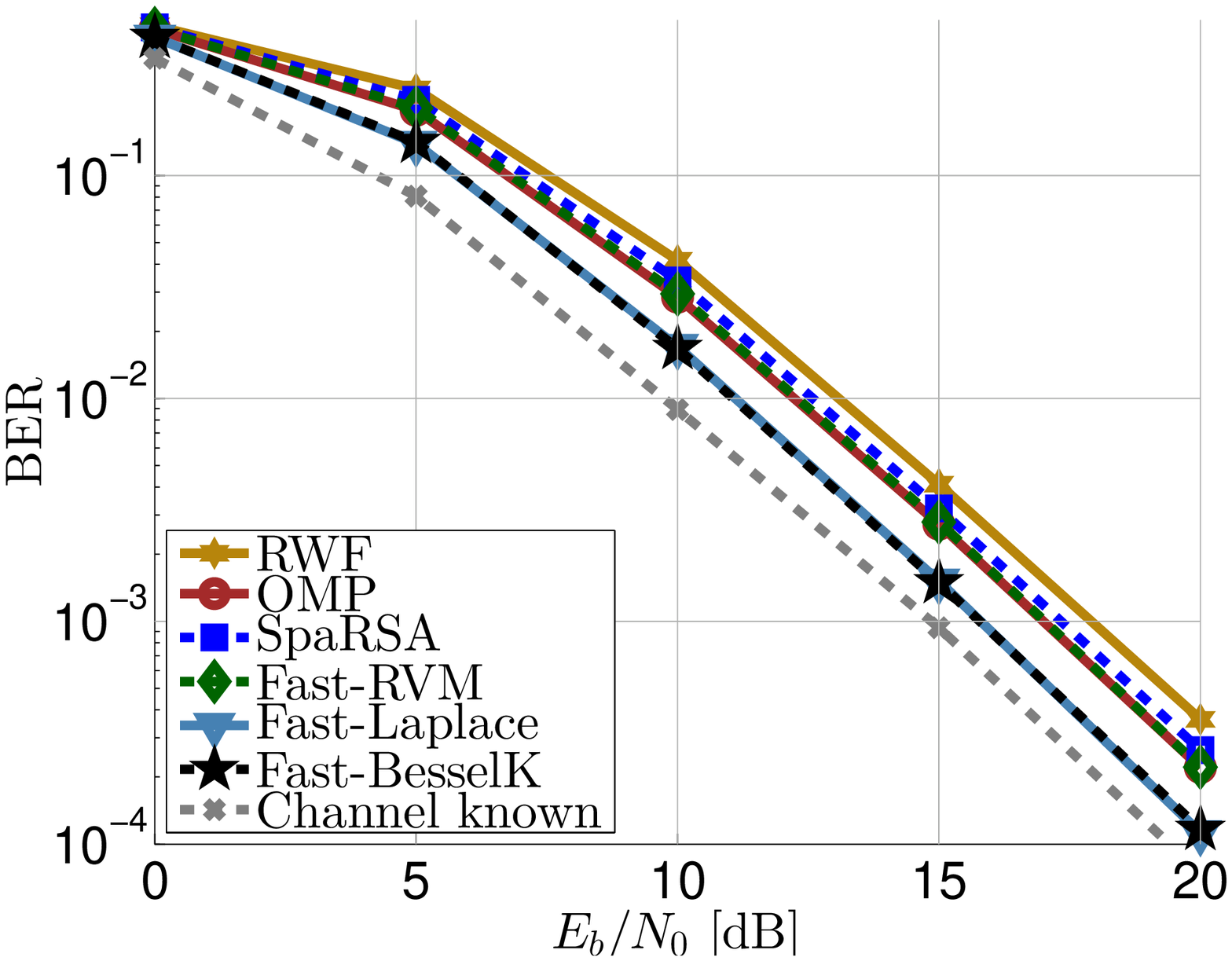}}
\subfigure[\label{fig:MSE_EbN0}]{\includegraphics[width=0.3\linewidth]{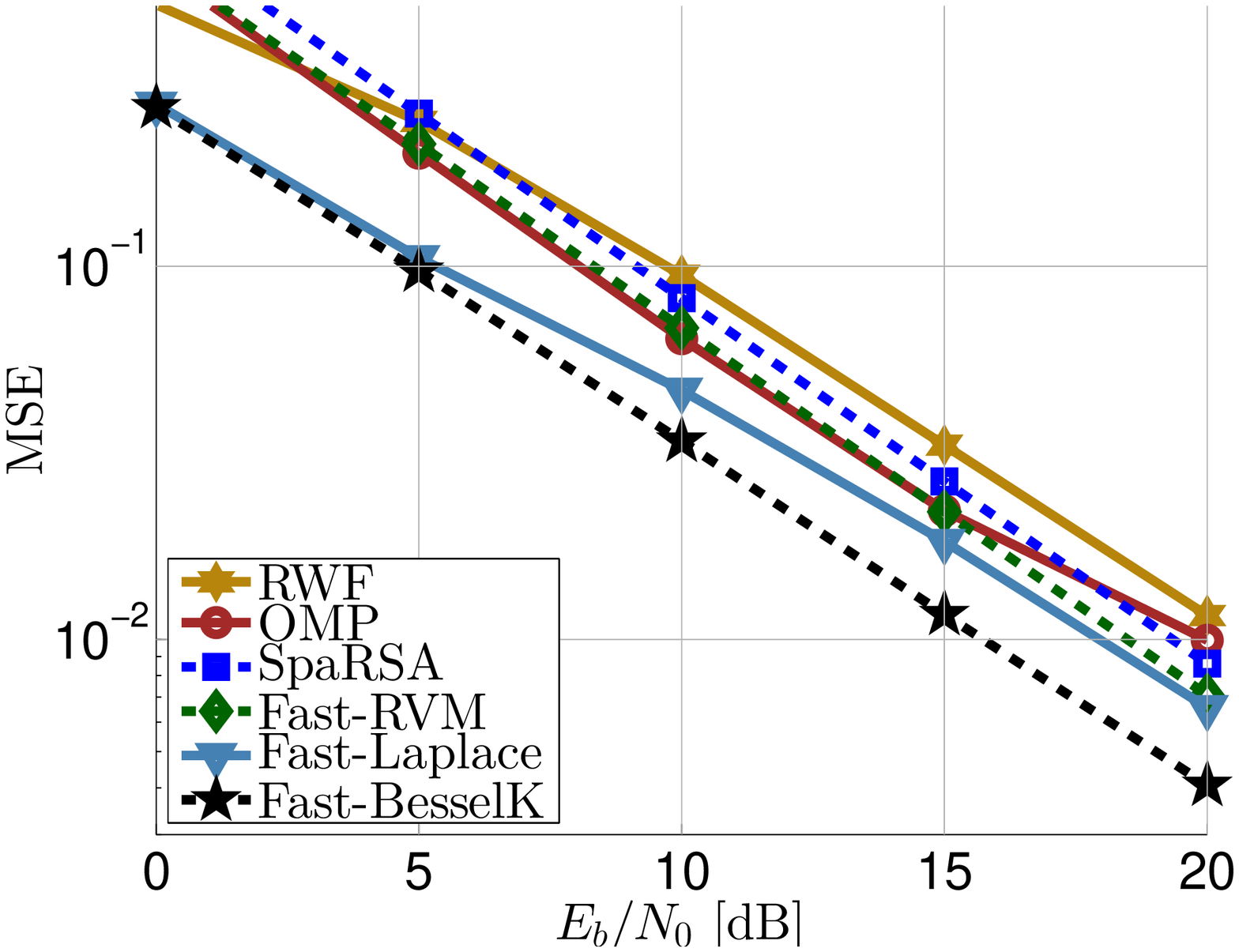}}
\subfigure[\label{fig:iter_EbN0}]{\includegraphics[width=0.3\linewidth]{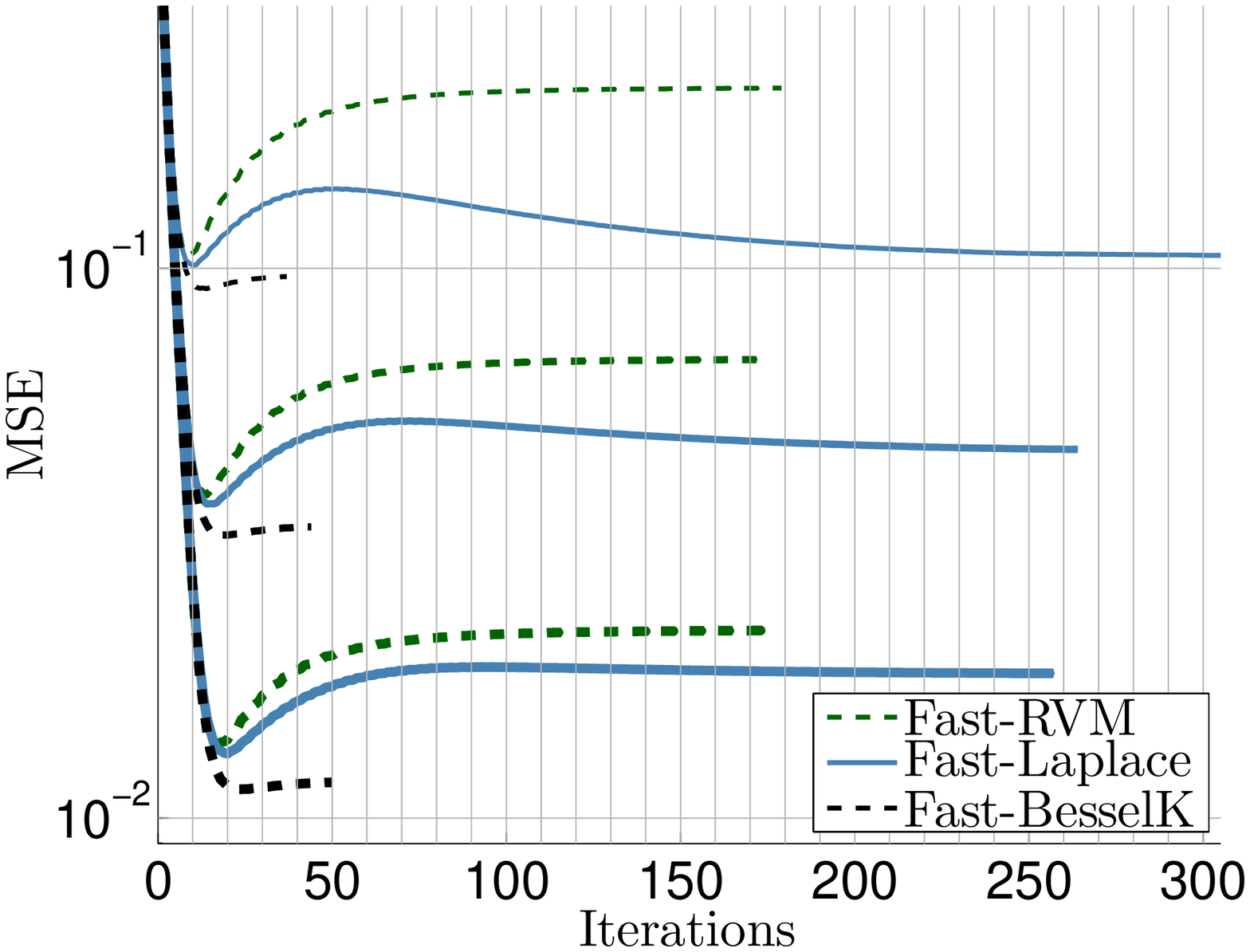}}
}
\caption{Performance comparison of the different algorithms: we have $M=100$, $L = 200$, and $\langle K \rangle = 10$. In (c) the SNR is fixed at 5 dB, 10 dB, and 15 dB.}
\label{fig:performance1}
\end{figure*} 
   
We follow the approach in \cite{Tipping2003} and realize the proposed fast iterative Bayesian inference algorithm by computing each $\hat{\gamma}_l$, $l=1,\ldots,L$, and selecting the one $\hat{\gamma}_l$ that gives rise to the greatest increase in $\ell(\hat{\gamma}_l)$ between two consecutive iterations. Depending on the new value $\hat{\gamma}_l$, we may then add, delete, or keep the corresponding column vector $\vect{\phi}_l$ in the dictionary. The quantities $\widehat{\matr{\Sigma}}$, $\hat{\vect{\mu}}$, and $\hat{\lambda}$ are updated using \eqref{eq:alpha_cov}, \eqref{eq:alpha_mu}, and \eqref{eq:lambda_mode} together with the computation of $s_l$ and $q_l$, $l=1,\ldots,L$. 
The computational complexity of each iteration is $O(LM\widehat{K})$ when $\widehat{K}<M<L$, where $\widehat{K}$ is the number of nonzero components in $\hat{\vect{\mu}}$. If $\hat{\lambda}$ is not updated between two consecutive iterations, $\widehat{\matr{\Sigma}}$, $\hat{\vect{\mu}}$, $s_l$, and $q_l$ can be updated efficiently according to the update procedures in \cite{Tipping2003}. In this case the cost in complexity is only $O(LM)$.
We refer to the proposed algorithm as \textit{Fast-BesselK}.

\subsection{Fast-RVM and Fast-Laplace}

The Fast-BesselK algorithm described in Section~\ref{sec:fast_besselk} is parametrized by $\epsilon$ and $\eta$. In the following, we will show how, by appropriately setting these parameters, we can obtain Fast-RVM \cite{Tipping2003} and Fast-Laplace \cite{Babacan2010} as particular instances of Fast-BesselK. For Fast-RVM, the estimation of $\gamma_l$ relies on the maximization of the likelihood $p(\vect{y}|\gamma_l,\hat{\vect{\gamma}}_{-l},\hat{\lambda})$, i.e., a constant prior is assumed for the hyperprior, $p(\gamma_l) \propto 1$. Hence, by selecting $\epsilon=1$ and $\eta = 0$ we obtain the cost function $\ell(\gamma_l)$ used in \cite{Tipping2003}. In case of Fast-Laplace \cite{Babacan2010}, the exponential pdf is selected for $p(\gamma_l)$. As the gamma pdf reduces to the exponential pdf by choosing its shape parameter $\epsilon=1$, we obtain $\ell(\gamma_l)$ used in \cite{Babacan2010} from this choice.



%% file: simulations.tex
\begin{figure*}[!t]
\centering
\centerline{
\subfigure[\label{fig:bv_EbN0}]{\includegraphics[width=0.24\linewidth]{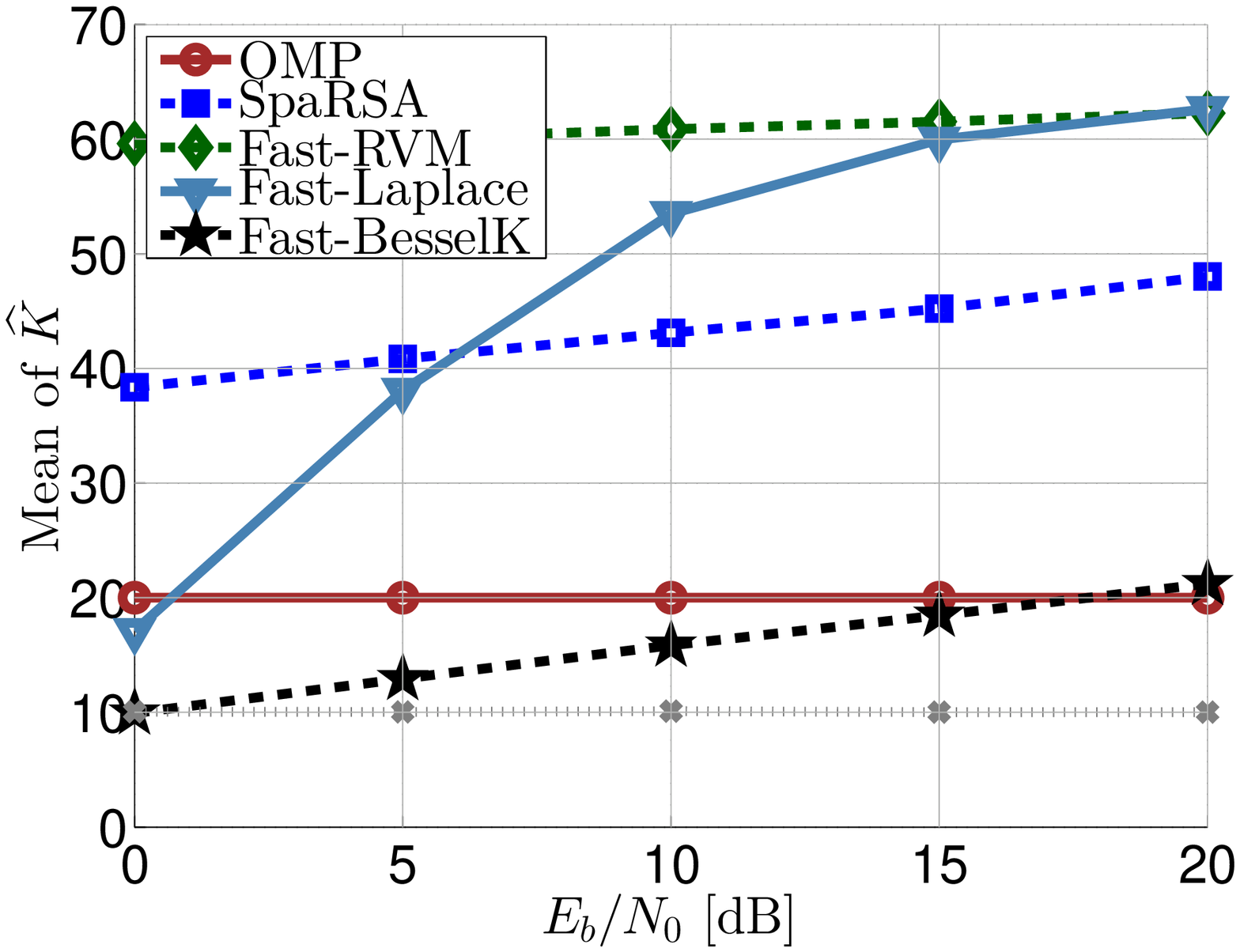}}
\subfigure[\label{fig:MSE_Delays}]{\includegraphics[width=0.24\linewidth]{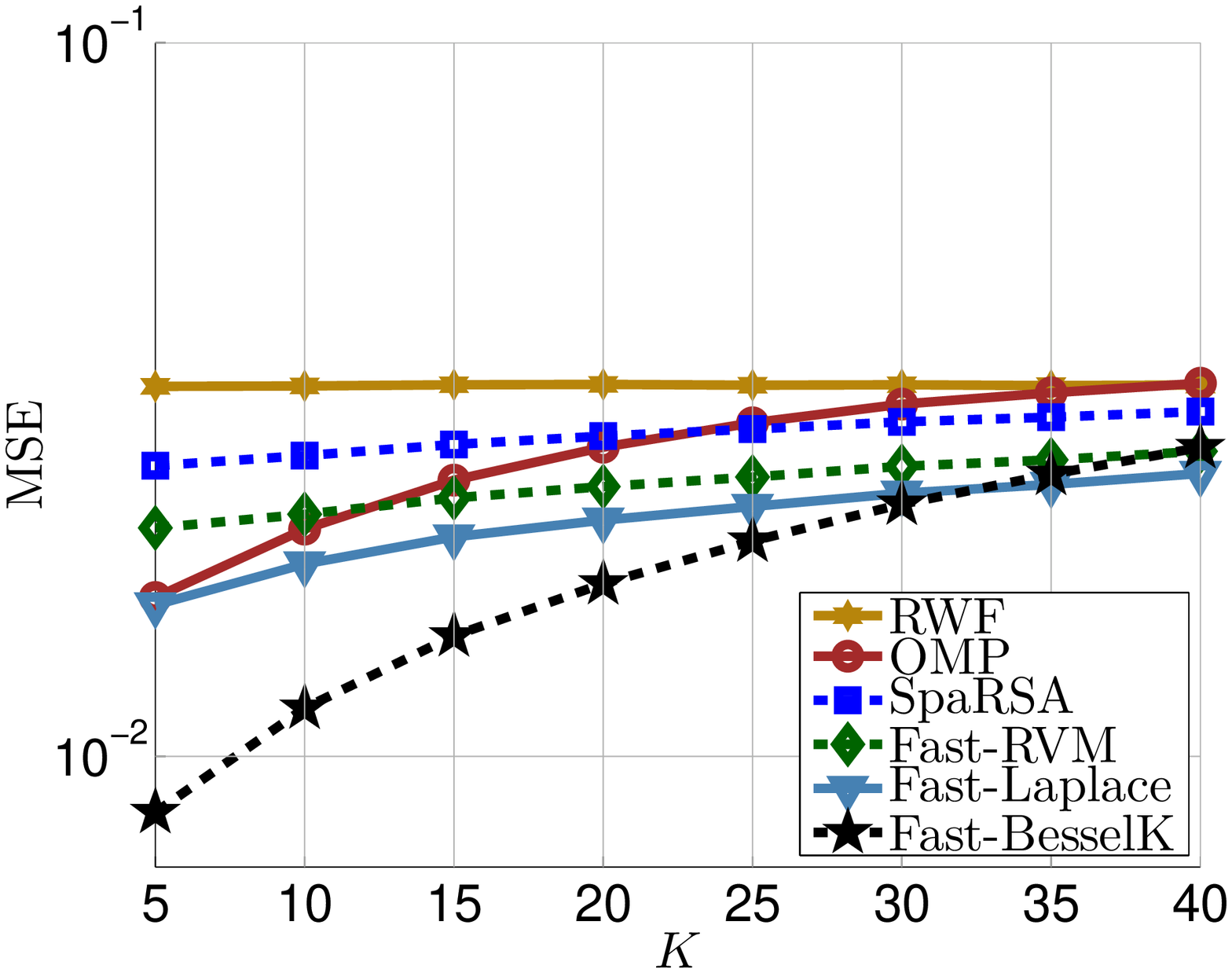}}
\subfigure[\label{fig:MSE_Pilots}]{\includegraphics[width=0.24\linewidth]{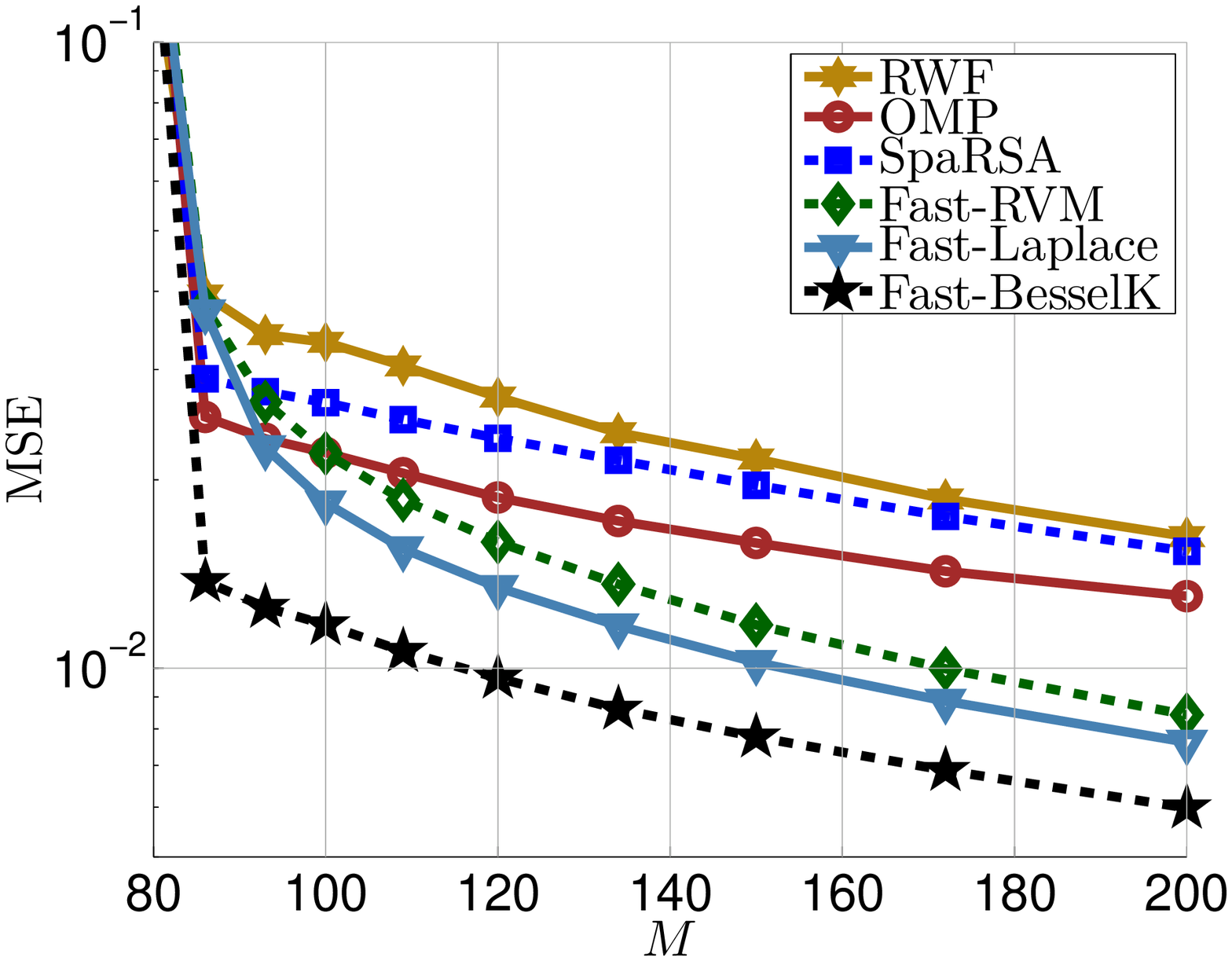}}
\subfigure[\label{fig:MSE_N}]{\includegraphics[width=0.24\linewidth]{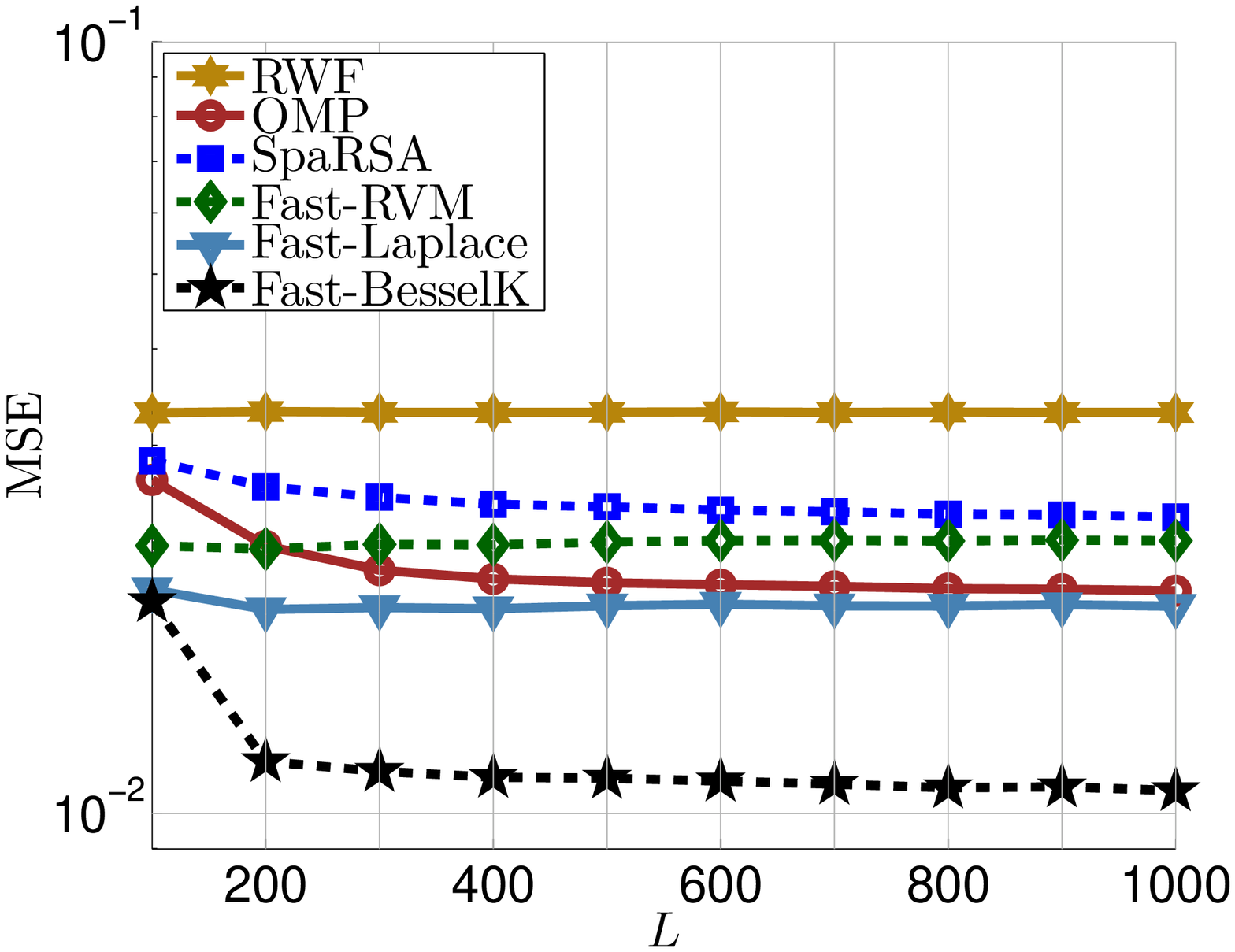}}
}
\caption{Performance comparison of the different algorithms: unless otherwise specified, $M=100$, $L = 200$, and $\langle K \rangle = 10$. In (b)-(d) the SNR is 15 dB. The dashed gray curve in (a) corresponds to $\langle K\rangle=10$.} 
\label{fig:performance2}
\end{figure*}

\begin{table}[!t]
	\centering
	\caption{Parameter settings for the simulations. \label{tab:ofdmsettings}}
	\vspace{-7pt} 
	\begin{tabular}{c|c}
	\hline
	Sampling time, $T_\mathrm{s}$ & 32.55 ns \\ 
	CP length & 4.69 $\mu$s / 144 $T_\mathrm{s}$\\
  Subcarrier spacing   & 15 kHz\\
	Pilot pattern & Evenly spaced, QPSK\\
	Modulation & QPSK ($M_d=2$) \\
	Subcarriers, $N$ & 1200 \\
	OFDM symbols & 1 \\
	Information bits & 1091 \\
	Channel interleaver & Random \\
	Convolutional code & $(133,171,165)_8$ \\
	Decoder & BCJR algorithm \cite{Bahl1974}
	\end{tabular}
\end{table}

We perform Monte Carlo simulations to evaluate the performance of Fast-BesselK derived in Section~\ref{sec:bil}. We consider a scenario inspired by the 3GPP LTE standard \cite{3GPP2008} with the settings specified in Table~\ref{tab:ofdmsettings}. In all investigations conducted we fix the spectral efficiency of $\kappa \triangleq M_d(N-M)R/N = 0.92$ information bits per subcarrier, which corresponds to a rate $R=1/2$ code. We note that we employ a rate-1/3 convolutional code and use puncturing in order to increase the spectral efficiency. Unless otherwise specified, $M=100$ evenly-spaced pilot symbols are used. 

The multipath channel \eqref{eq:channel} is based on the model used in \cite{Jakobsen2010} where, for each realization of the channel, the total number of multipath components $K$ is Poisson distributed with mean $\langle K \rangle = 10$ and the delays $\tau_k$, $k =1,\ldots,K$, are independent and uniformly distributed random variables drawn from the continuous interval $[0,144~T_\mathrm{s}]$. Conditioned on $\tau_k$, $k =1,\ldots,K$, the weights  $\beta_k$, $k =1,\ldots,K$, are independent, and weight $\beta_k$ has a zero-mean complex circular symmetric Gaussian distribution with variance $\sigma^2(\tau_k) = u\exp(-\tau_k/v)$ and parameters $u,v > 0$.\footnote{The parameter $u$ is computed such that $\langle \sum_{k=1}^{K} |\beta_k|^2 \rangle = 1$. In the considered simulation scenario, $\langle K\rangle = 10$, $\tau_\textrm{max} = 144\; T_\mathrm{s}$, and $v = 40\; T_\mathrm{s}$.} In this way $\{\tau_k,\beta_k\}$ form a marked Poisson process. 

For Fast-BesselK, we set $\epsilon=0.5$ and $\eta=1$ in all investigations. We empirically observed that this is a proper selection of parameters for channel models with both few and numerous multipath components. Fast-BesselK is compared to two Bayesian methods, Fast-RVM \cite{Tipping2003}\footnote{The software is available at \url{http://people.ee.duke.edu/~lcarin/BCS.html}.} and Fast-Laplace \cite{Babacan2010}\footnote{The software is available at \url{http://ivpl.eecs.northwestern.edu/}.}. For these three algorithms the noise precision $\lambda$ is estimated at every third iteration with the initialization $\Var(\vect{y})/100$ \cite{Tipping2003}. The stopping criterion is based on the difference in $\ell(\hat{\gamma_l})$ between two consecutive iterations \cite{Ji2008}. Two non-Bayesian methods, LASSO and OMP, are also included for comparison. For LASSO, we use the sparse reconstruction by separable approximation (SpaRSA) algorithm \cite{Wright2009}\footnote{The software is available on-line at \url{http://www.lx.it.pt/~mtf/SpaRSA/}}. The required regularization parameter is chosen as $5\sqrt{\log(L)/\lambda}$ \cite{Ben-Haim2009}, which has been empirically observed to provide satisfactory results. For OMP, an a priori estimate of the sparsity of $\vect{\alpha}$ needs to be set. In all investigations we use $\langle K \rangle+10$. 
Finally, the commonly employed robustly designed Wiener filter (RWF) \cite{Edfors1998} for OFDM channel estimation is used as a reference.


Unless otherwise specified, we set the number of rows in $\matr{\Phi}$ to $M=100$ (pilot subcarriers) and the number of columns in $\matr{\Phi}$ to $L=200$, which corresponds to a delay resolution of $T_s/\zeta = 0.72$ $T_\mathrm{s}$. The performance versus SNR is compared in Figs.~\ref{fig:BER_EbN0}-\ref{fig:MSE_EbN0}. From Fig.~\ref{fig:BER_EbN0}, we see that Fast-BesselK and Fast-Laplace outperform the other algorithms in terms of BER across all the SNR range considered. Specifically, at 1 $\%$ BER the gain is apporiximatly 1 dB over Fast-RVM, LASSO, and OMP and 2 dB over RWF. Fig.~\ref{fig:MSE_EbN0} shows how Fast-BesselK yields a lower MSE than the other algorithms. Surprisingly, the improved performance in MSE achieved by Fast-BesselK does not lead to a better BER performance when compared to Fast-Laplace. 

The convergence speed of the Bayesian iterative algorithms is shown in Fig.~\ref{fig:iter_EbN0}. Here, Fast-BesselK achieves a remarkable improvement compared to Fast-RVM and Fast-Laplace with MSE values converging in about 10-30 iterations. As Fig.~\ref{fig:iter_EbN0} shows, there is no guarantee that the MSE is reduced at each iteration, due to the objective function \eqref{eq:sbl_cost}. Fast-RVM and Fast-Laplace suffer a significant increase in MSE after a certain number of iterations; this drawback is significantly mitigated in the case of Fast-BesselK. The superior convergence speed of Fast-BesselK can be explained by observing Figs.~\ref{fig:bv_EbN0}-\ref{fig:MSE_Delays}.  
Fig.~\ref{fig:MSE_Delays} shows that the improvement in convergence rate comes as the Besssel K prior can handle channels with few multipath components better (i.e., yields lower MSE). As a consequence, the other methods tend to add more column vectors to the dictionary matrix, thus, increasing the number of add, delete, and reestimate iterations as seen from Fig.~\ref{fig:bv_EbN0}. 

Fig.~\ref{fig:MSE_Pilots} shows the MSE versus the number of pilots $M$. We observe that, for a given MSE performance, Fast-BesselK is able to significantly reduce the required pilot overhead. In particular, Fast-BesselK achieves an MSE on pair with LASSO, OMP, and RWF using less than half the number of pilots. Finally, in Fig.~\ref{fig:MSE_N} we evaluate the MSE performance versus available delay resolution determined by the number of columns $L$ in $\matr{\Phi}$ (cf., Section~\ref{sec:system}).\footnote{Naturally, RWF does not require a dictionary matrix $\matr{\Phi}$ to be specified and its performance is thereby independent of $L$.} Several observations are worth being noticed. Fast-BesselK leads to a noticeable MSE performance gain as the delay resolution improves as opposed to the other algorithms. In fact, it appears that, besides Fast-BesselK, only OMP is able to exploit the improved delay resolution. The reason for this is that LASSO, Fast-RVM, and Fast-Laplace produce a solution $\hvect{h}_\ipilot = \matr{\Phi}\hvect{\alpha}$ with an increasing number of nonzero components $\widehat{K}$ in $\hvect{\alpha}$ when increasing $L$ (there are simply more column vectors in $\matr{\Phi}$ to be added or deleted). Thus, these algorithms also require an increasing amount of iterations to be run as opposed to Fast-BesselK (results not shown).



%% file: conclusion.tex
In this work, we presented a fast iterative Bayesian inference channel estimation algorithm based on the hierarchical Bayesian prior model of the Bessel K probability density function. Following the framework for fast Bayesian inference in \cite{Tipping2003}, we proposed an algorithm that significantly lowers the number of needed iterations as compared to state-of-the-art Bayesian inference methods with no penalization in performance. This improvement in convergence rate is directly related to the Bessel K prior's ability to handle channels with few multipath components better than other commonly employed prior models. Furthermore, our algorithm shows improved performance when compared to both Bayesian and non-Bayesian state-of-the-art methods.